\documentclass[11pt]{article}

\usepackage[preprint]{acl}
\usepackage{times}
\usepackage{latexsym}
\usepackage[T1]{fontenc}
\usepackage[utf8]{inputenc}
\usepackage{microtype}
\usepackage{inconsolata}
\usepackage{graphicx}
\usepackage{booktabs}
\usepackage{multirow}
\usepackage{comment}
\hypersetup{colorlinks=true, linkcolor=blue, citecolor=blue, urlcolor=blue}
\title{Investigating Spatial Attention Bias in Vision-Language Models}

\author{
  Aryan Chaudhary$^1$, Sanchit Goyal$^1$, Pratik Narang$^1$, Dhruv Kumar$^1$ \\
  $^1$Birla Institute of Technology and Science, Pilani, India \\
  \textbf{Correspondence:} \href{mailto:f20230651@pilani.bits-pilani.ac.in}{f20230651@pilani.bits-pilani.ac.in}
}
\begin{document}
\maketitle
\begin{abstract}
Vision-Language Models have demonstrated remarkable capabilities in understanding visual content, yet systematic biases in their spatial processing remain largely unexplored. This work identifies and characterizes a systematic spatial attention bias where VLMs consistently prioritize describing left-positioned content before right-positioned content in horizontally concatenated images. Through controlled experiments on image pairs using both open-source and closed-source models, we demonstrate that this bias persists across different architectures, with models describing left-positioned content first in approximately 97\% of cases under neutral prompting conditions. Testing on an Arabic-finetuned model reveals that the bias persists despite right-to-left language training, ruling out language reading direction as the primary cause. Investigation of training dataset annotation guidelines from PixMo and Visual Genome reveals no explicit left-first ordering instructions, suggesting the bias is consistent with architectural factors rather than explicit training data instructions. These findings reveal fundamental limitations in how current VLMs process spatial information.
\end{abstract}
\section{Introduction}

Vision-Language Models (VLMs) have achieved significant advances in multimodal understanding by jointly modeling visual and textual modalities, enabling applications such as image captioning, visual question answering, and multimodal reasoning~\cite{krishna2017visual,lu2019vilbert,li2020oscar,radford2021clip}. Large-scale pretrained VLMs now serve as general-purpose multimodal systems and exhibit strong zero-shot and in-context learning capabilities~\cite{alayrac2022flamingo,li2023blip2}.

Despite these achievements, systematic biases in how VLMs process spatial information remain poorly understood~\cite{zhu2025bias}. Such biases can compromise model reliability in real-world applications, particularly those requiring fair and balanced visual interpretation, including user interfaces, robotics, and decision-support systems.

This work investigates a specific manifestation of spatial processing bias: when presented with horizontally concatenated images containing two distinct objects, do VLMs exhibit systematic preferences in the order they describe spatial content? We focus on two fundamental questions. First, do VLMs demonstrate consistent spatial attention bias across different architectures and model generations? Second, what factors contribute to this bias. Does it stem from language reading direction, training data annotation patterns, or architectural design choices?

We conduct experiments using a controlled dataset of 100 image pairs from Caltech-101~\cite{fei2004learning}, where 50 unique pairs are presented alongside their horizontal flips to isolate positional effects from content-specific biases. We evaluate seven models spanning both open-source architectures including InternVL2-8B and InternVL3-8B~\cite{chen2024internvl,internvl2hf,internvl3hf}, Qwen2-VL-7B~\cite{qwen2024qwen2vl,qwen2vlhf}, and Qwen2.5-VL-7B~\cite{bai2025qwen25vl} and closed-source systems, including GPT-5 Nano~\cite{openai2025gpt5}, Gemini 2.0 Flash~\cite{gemini2flash_modelcard}, and Claude 4.5 Haiku~\cite{anthropic2025claude45}. Each model is tested with both simple directional prompts and structured JSON-format prompts to assess natural behavior and responsiveness to explicit instructions.The key contributions of this work are as \mbox{follows}:
\begin{itemize}
    \item We provide a systematic characterization of spatial attention bias across seven diverse VLM architectures, identifying a consistent left-to-right processing preference in open-source models.
    \item We demonstrate through Arabic-finetuned model testing~\cite{oryx2025ain} that this bias persists regardless of language reading direction (right-to-left), ruling out language priors as the primary cause.
    \item We analyze annotation guidelines from PixMo~\cite{deitke2024molmo} and Visual Genome~\cite{krishna2017visual}, showing that explicit left-first ordering instructions are absent from major training datasets.
    \item We evaluate bias on a \emph{Desktop UI} dataset consisting of dense interface elements, highlighting how spatial preferences shift in information-rich contexts.
\end{itemize}

\section{Related Work}
\begin{figure*}[t!]
    \centering
    \includegraphics[width=\linewidth]{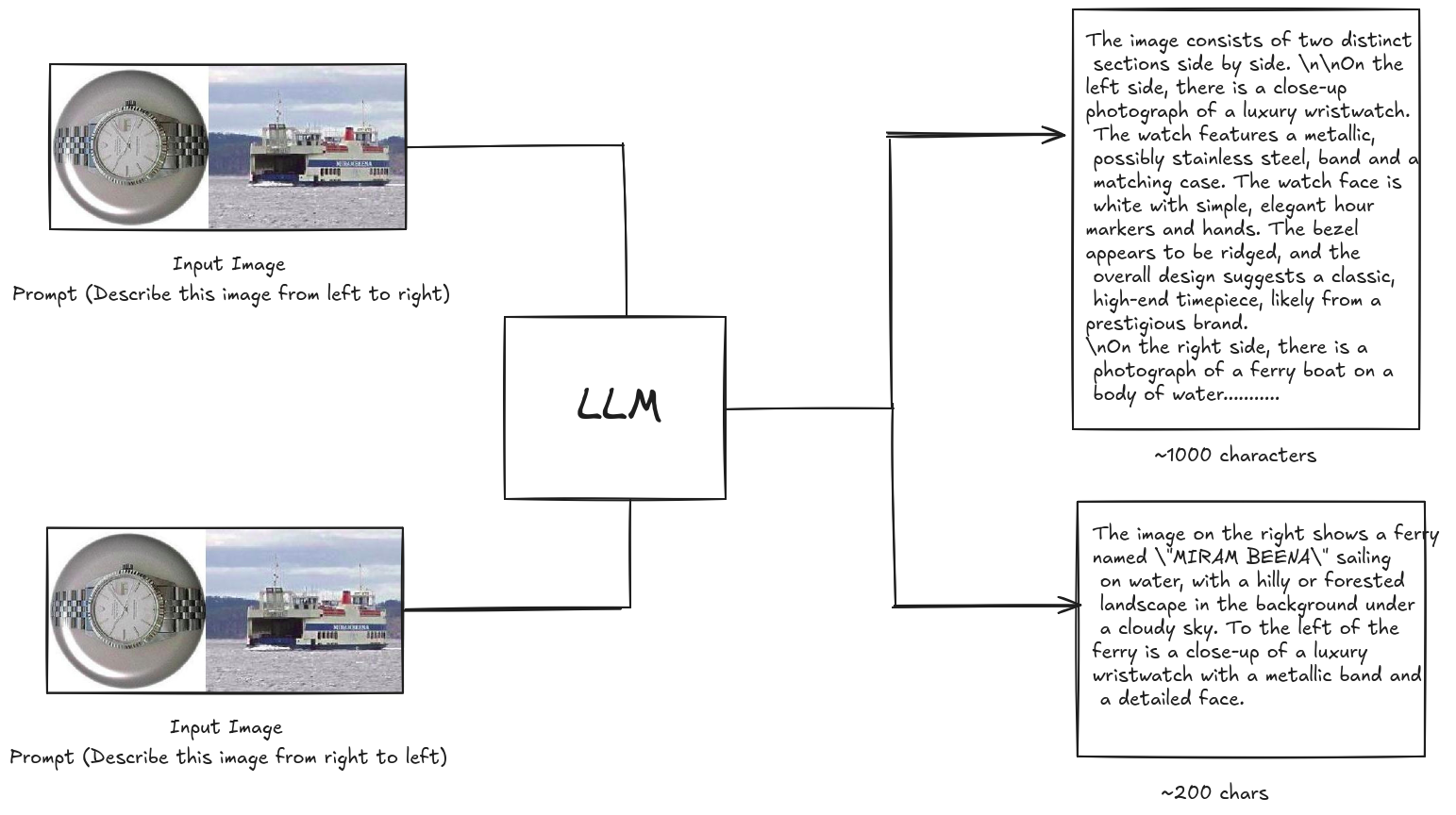}
    \caption{Overview of the experimental setup for detecting spatial attention bias.}
    \label{fig:llm_overview}
\end{figure*}
\subsection{Vision-Language Models}
  Vision-Language Models (VLMs) typically feature a modular architecture comprising a frozen visual encoder \cite{radford2021learning, zhai2023sigmoid}, a large language model backbone \cite{touvron2023llama, chiang2023vicuna}, and a cross-modal alignment projector \cite{alayrac2022flamingo, li2023blip2, liu2024llava} that bridges the semantic gap between modalities. These systems are initially grounded on massive image-text corpora \cite{schuhmann2022laion} and subsequently refined through visual instruction tuning \cite{ dai2023instructblip} or direct preference optimization \cite{zhao2024silky, yu2024rlhf} to enhance instruction adherence. This paradigm has yielded remarkable proficiency across diverse applications, including open-ended dialogue \cite{bai2023qwen, chen2024internvl}, complex visual reasoning \cite{wang2023cogvlm, zhu2023minigpt4}, and domain-specific tasks such as biomedical analysis \cite{li2024llavamed}. However, despite these advancements, recent research suggests that mapping high-dimensional visual features into the causal attention mechanisms of LLMs introduces structural vulnerabilities, particularly regarding how spatial information is encoded, preserved, and prioritized during cross-modal interactions.

\subsection{Biases in Multimodal Models}
Recent work has identified various biases in VLMs. Gender bias in image captioning has been extensively documented, with models often making stereotypical associations based on statistical correlations in training data \cite{hendricks2018women, zhao2021understanding}. Object co-occurrence biases are also prevalent, causing models to hallucinate commonly associated objects even when the object is absent from the image \cite{rohrbach2018object, li2023pope}. While these semantic biases have been widely studied, systematic biases in spatial processing remain comparatively underexplored.

\subsection{Spatial and Positional Bias}
Recent investigations have begun to explore the spatial robustness of LVLMs, though the focus has largely remained on reasoning accuracy rather than scanning priority. Notably, \citet{zhu2025bias} conducted a systematic study demonstrating that current LVLMs often produce inconsistent semantic outputs when identical visual content is shifted to different spatial locations. More recently, \citet{tian2025identifying} investigated position bias in multi-image contexts, identifying a "recency bias" in open-source models (preferring the last image) and a "lost-in-the-middle" phenomenon in proprietary models when performing specific reasoning tasks.

While these studies highlight vulnerabilities in performance stability (specifically whether a model answers correctly depending on image placement), they do not address the distinct phenomenon of \textit{sequential attention bias}, or the intrinsic preference to describe specific spatial regions first. Furthermore, existing literature has not sufficiently isolated visual scanning biases from linguistic reading directionality, nor evaluated how these spatial priors manifest in information-dense environments like user interfaces. Our work addresses these gaps by decoupling text generation order from visual processing and examining how entrenched spatial habits can lead to instruction-following failures in state-of-the-art models.
\section{Methodology}
\label{sec:method}
\subsection{Dataset Construction}

To evaluate spatial bias across different levels of visual complexity and information density, we constructed two distinct datasets: one focusing on object-centric natural images and another on dense, text-rich user interfaces.

\subsubsection{Caltech-101 Controlled Pairs} We constructed a controlled dataset to isolate positional bias from content-specific effects. From Caltech-101 \cite{fei2004learning}, we selected 50 diverse image pairs from different object categories. For each pair, we created the original left-right arrangement and a horizontally flipped version, yielding 100 total test images. This design ensures that observed bias cannot be attributed to specific image content being inherently more description-worthy. All images were resized to 512×512 pixels using bicubic interpolation before horizontal concatenation.

\subsubsection{Desktop UI Dataset} To investigate model behavior in high-density information contexts, we curated a "Desktop UI" dataset comprising 10 distinct screenshots of common digital interfaces (e.g., LinkedIn feed, Google News, Twitter/X feed, YouTube homepage, Gmail inbox, and LeetCode). From these 10 base screenshots, we generated all unordered pairwise combinations (10C2 = 45 unique pairs), and horizontally concatenated each pair to form the evaluation set. This design allows us to systematically evaluate spatial bias across diverse UI layouts while controlling for the underlying content. Unlike the object-centric Caltech dataset, these images contain dense textual information and complex UI elements, testing whether spatial scanning habits shift when models are presented with reading-heavy tasks.

\subsection{Experimental Design}

We conducted three sets of experiments to investigate different aspects of spatial bias.

\subsubsection{Neutral Prompting Experiment}

To establish baseline spatial preference without explicit directional instructions, we tested Qwen2.5-VL-7B and the Arabic-finetuned AIN model using the neutral prompt "Describe the image" with no directional guidance. Responses were analyzed to determine which image (left or right) was mentioned first in the description.

\subsubsection{Directional Prompting Experiments}

Models were evaluated under two prompting conditions. Simple prompts used "Describe the image from left to right" and "Describe the image from right to left." Structured prompts used "Describe the image from left to right in structured JSON format, Left image: <Description> Right Image: <Description>" and the corresponding right-to-left variant. These conditions test both natural behavior and responsiveness to explicit format constraints.

\subsubsection{Dataset Annotation Analysis}

We examined annotation guidelines from two major vision-language training datasets: PixMo, which uses high-quality human annotations collected through speech-based descriptions, and Visual Genome, a widely-used dataset for vision-language pretraining. This analysis aimed to determine whether explicit left-to-right ordering instructions exist in dataset creation protocols.

\subsection{Models Evaluated}

We selected seven models representing diverse architectures and training approaches. Open-source models include InternVL2-8B \cite{internvl2hf}, InternVL3-8B \cite{internvl3hf}, Qwen2-VL-7B \cite{qwen2024qwen2vl,qwen2vlhf}, and Qwen2.5-VL-7B \cite{bai2025qwen25vl}. Closed-source systems include GPT-5 Nano \cite{openai2025gpt5}, Gemini 2.0 Flash \cite{gemini2flash_modelcard}, and Claude 4.5 Haiku \cite{anthropic2025claude45}. Additionally, we tested the AIN model fine-tuned for Arabic \cite{oryx2025ain} to examine whether language reading direction influences spatial processing bias.

All models were run using consistent inference configuration with temperature set to 0.5 and top-p sampling at 0.9 to balance generation diversity with consistency.

\subsection{Evaluation Metrics}

For the neutral prompting experiment, we determined which spatial position was mentioned first in model responses and calculated the percentage of left-first descriptions.

For the Directional Prompting Experiments we measured bias through description length analysis, calculating the difference between response lengths for left-to-right versus right-to-left prompting. The metric is defined as the length of the left-to-right response minus the length of the right-to-left response. Positive values indicate longer descriptions when prompted to describe from left to right, suggesting models allocate more descriptive attention to left-positioned content even when explicitly instructed to prioritize right-positioned content.

We established a threshold of ±60 characters to identify cases of substantial bias while tracking the full distribution of differences. For each model and prompting condition, we calculated the percentage of cases exceeding +60 character difference (strong left bias), percentage below -60 characters (strong right bias), percentage within ±60 characters (relatively balanced), average difference across all test cases, and maximum and minimum observed differences.

\section{Results}
\subsection{Neutral Prompting Baseline}
\label{sec:neutral_results}

Before evaluating the magnitude of bias using character counts, we first established the baseline scanning preference of VLMs using a neutral prompt. We evaluated \textbf{Qwen2.5-VL-7B} \cite{bai2025qwen25vl} and the Arabic-finetuned \textbf{AIN} model \cite{oryx2025ain} on the 100-pair Caltech-101 dataset.

Both models were given the instruction \textit{"Describe the image"} with no directional constraints or formatting cues. We manually analyzed the responses to identify which image in the pair was described first.
\begin{table}[h]
\centering
\small
\begin{tabular}{@{}llc@{}}
\toprule
\textbf{Model} & \textbf{Training Prior} & \textbf{Left-First Preference} \\
\midrule
Qwen2.5-VL-7B & English (LTR) & \textbf{97\%} \\
AIN (Arabic) & Arabic (RTL) & \textbf{97\%} \\
\bottomrule
\end{tabular}
\caption{Results of the \textbf{Neutral Prompting Experiment}. Despite the AIN model being fine-tuned on Right-to-Left (RTL) Arabic data, it exhibits an identical Left-to-Right scanning bias to the English-trained Qwen2.5-VL.}
\label{tab:neutral_bias}
\end{table}

The results, presented in Table~\ref{tab:neutral_bias}, reveal a near-universal preference for left-to-right processing. Both models described the left image first in \textbf{97\%} of cases. This finding is critical because the AIN model is specifically fine-tuned on Arabic instructions and data, which naturally follows a Right-to-Left (RTL) reading order. If the spatial bias were driven by the language model's learned text generation direction, the Arabic model should favor a Right-to-Left description order. The fact that it instead adheres to a Left-to-Right visual scan strongly suggests that this bias is intrinsic to the visual encoder or the cross-modal alignment process, operating independently of the language decoder's textual training.

\subsection{Spatial Bias in Object-Centric Images}

Our initial experiments on the Caltech-101 dataset reveal significant spatial bias across all tested models, though with considerable variation in magnitude. Table~\ref{tab:caltech_summary} presents the key findings for object-centric image pairs.

\begin{table*}[t]
\centering
\small
\setlength{\tabcolsep}{5pt}
\begin{tabular}{@{}lcccccc@{}}
\toprule
 & \multicolumn{3}{c}{\textbf{Bias Frequency (\%)}} & \multicolumn{3}{c}{\textbf{Character Difference Magnitude}} \\
\cmidrule(lr){2-4} \cmidrule(l){5-7}
\textbf{Model} & \textbf{Left Bias} & \textbf{Balanced} & \textbf{Right Bias} & \textbf{Avg Diff} & \textbf{Max} & \textbf{Min} \\
 & \textit{(>+60)} & \textit{(Within $\pm$60)} & \textit{(<-60)} & & & \\
\midrule
InternVL2-8B & 80 & 12 & 8 & +472.8 & +1619 & -600 \\
InternVL3-8B & 77 & 18 & 5 & +393.5 & +1330 & -242 \\
Qwen2-VL-7B & 39 & 47 & 14 & +81.1 & +1194 & -1538 \\
Qwen2.5-VL-7B & 48 & 19 & 33 & +54.3 & +794 & -374 \\
\midrule
GPT-5 Nano & 14 & 77 & 9 & +3.3 & +186 & -116 \\
Gemini 2.0 Flash & 37 & 47 & 16 & +43.0 & +345 & -249 \\
Claude 4.5 Haiku & 39 & 30 & 31 & +14.0 & +392 & -263 \\
\bottomrule
\end{tabular}
\caption{Summary of spatial bias on the \textbf{Caltech-101 dataset} under \textbf{Simple Prompt}. \textbf{Left Bias} indicates the percentage of cases where the Left-to-Right description was significantly longer (>60 chars) than the Right-to-Left description. \textbf{Balanced} indicates cases with negligible difference ($\pm$60 chars).}
\label{tab:caltech_summary}
\end{table*}

\begin{table*}[t]
\centering
\small
\setlength{\tabcolsep}{5pt}
\begin{tabular}{@{}lcccccc@{}}
\toprule
 & \multicolumn{3}{c}{\textbf{Bias Frequency (\%)}} & \multicolumn{3}{c}{\textbf{Character Difference Magnitude}} \\
\cmidrule(lr){2-4} \cmidrule(l){5-7}
\textbf{Model} & \textbf{Left Bias} & \textbf{Balanced} & \textbf{Right Bias} & \textbf{Avg Diff} & \textbf{Max} & \textbf{Min} \\
 & \textit{(>+60)} & \textit{(Within $\pm$60)} & \textit{(<-60)} & & & \\
\midrule
InternVL2-8B & 47 & 24 & 29 & +33.8 & +445 & -413 \\
InternVL3-8B & 44 & 40 & 16 & +41.0 & +296 & -238 \\
Qwen2.5-VL-7B & 45 & 25 & 30 & +41.4 & +632 & -505 \\
\midrule
GPT-5 Nano & 11 & 75 & 14 & -1.9 & +137 & -139 \\
Gemini 2.0 Flash & 20 & 44 & 36 & -30.2 & +377 & -290 \\
Claude 4.5 Haiku & 35 & 37 & 28 & +11.4 & +441 & -312 \\
\bottomrule
\end{tabular}
\caption{Summary of spatial bias on the \textbf{Caltech-101 dataset} under \textbf{Structured Prompting} (JSON format). Explicit formatting instructions increase the frequency of Balanced responses, particularly in closed-source models.}
\label{tab:caltech_structured}
\end{table*}

Open-source models demonstrate a pronounced left-to-right bias. InternVL2-8B exhibits the strongest effect, with an average difference of +472.8 characters, indicating substantially longer descriptions when following the natural left-to-right tendency. InternVL3-8B shows similar patterns (+393.5 characters), suggesting architectural improvements did not resolve this spatial preference.

Closed-source models generally show better balance on object-centric data. GPT-5 Nano achieves near-zero average difference (+3.3), and Claude 4.5 Haiku shows only a modest preference (+14.0). However, individual variance remains high, indicating that while the aggregate behavior is balanced, specific instances still trigger directional preferences.

\subsection{Spatial Bias in Dense UI Contexts}

We extended our evaluation to the "Desktop UI" dataset to determine if spatial bias persists or shifts in dense, text-rich UI contexts. The results, summarized in Table~\ref{tab:computer_use_summary}, indicate that high information density significantly alters scanning behaviors for certain models.

\begin{table*}[t]
\centering
\small
\setlength{\tabcolsep}{5pt}
\begin{tabular}{@{}lcccccc@{}}
\toprule
 & \multicolumn{3}{c}{\textbf{Bias Frequency (\%)}} & \multicolumn{3}{c}{\textbf{Character Difference Magnitude}} \\
\cmidrule(lr){2-4} \cmidrule(l){5-7}
\textbf{Model} & \textbf{Left Bias} & \textbf{Balanced} & \textbf{Right Bias} & \textbf{Avg Diff} & \textbf{Max} & \textbf{Min} \\
 & \textit{(>+60)} & \textit{(Within $\pm$60)} & \textit{(<-60)} & & & \\
\midrule
InternVL2-8B & 42 & 27 & 31 & +30.7 & +344 & -301 \\
InternVL3-8B & 60 & 7 & 33 & +114.6 & +886 & -761 \\
Qwen2.5-VL-7B & 36 & 30 & 34 & -7.6 & +735 & -1181 \\
\midrule
GPT-5 Nano & 24 & 9 & 67 & -179.7 & +463 & -776 \\
Gemini 2.0 Flash & 43 & 5 & 52 & -33.9 & +1504 & -1651 \\
Claude 4.5 Haiku & 44 & 33 & 23 & +45.7 & +355 & -293 \\
\bottomrule
\end{tabular}
\caption{Summary of spatial bias on the \textbf{Desktop UI dataset} under \textbf{Simple Prompt}. High information density triggers a shift toward Right Bias (negative Average Difference) in models like GPT-5 Nano and Gemini 2.0 Flash.}
\label{tab:computer_use_summary}
\end{table*}
\begin{table*}[t]
\centering
\small
\setlength{\tabcolsep}{5pt}
\begin{tabular}{@{}lcccccc@{}}
\toprule
 & \multicolumn{3}{c}{\textbf{Bias Frequency (\%)}} & \multicolumn{3}{c}{\textbf{Character Difference Magnitude}} \\
\cmidrule(lr){2-4} \cmidrule(l){5-7}
\textbf{Model} & \textbf{Left Bias} & \textbf{Balanced} & \textbf{Right Bias} & \textbf{Avg Diff} & \textbf{Max} & \textbf{Min} \\
 & \textit{(>+60)} & \textit{(Within $\pm$60)} & \textit{(<-60)} & & & \\
\midrule
InternVL2-8B & 44 & 16 & 40 & -17.4 & +1558 & -2032 \\
InternVL3-8B & 33 & 16 & 51 & -75.0 & +812 & -1932 \\
\midrule
GPT-5 Nano & 41 & 15 & 44 & +11.2 & +711 & -839 \\
Gemini 2.0 Flash & 34 & 9 & 57 & -115.9 & +1759 & -1094 \\
Claude 4.5 Haiku & 46 & 21 & 33 & +41.0 & +556 & -683 \\
\bottomrule
\end{tabular}
\caption{Summary of spatial bias on the \textbf{Desktop UI dataset} under \textbf{Structured Prompting}. In dense UI contexts, structured formatting fails to enforce balance, with models like Gemini 2.0 Flash and InternVL3 exhibiting significant Right Bias.}
\label{tab:desktop_structured}
\end{table*}

\paragraph{Shift in Polarity}
Unlike the consistent left-bias seen in object recognition, the Desktop UI dataset induced a strong right-favoring bias (negative average difference) in several capable models. GPT-5 Nano shifted from a low bias on Caltech to a significant -179.7 on Dense UI data, with 67\% of cases favoring the right image. Similarly, Gemini 2.0 Flash shifted to -33.9. This suggests that when presented with dense text or UI feeds, these models may allocate more attention to the latter half of the context window or the right side of the visual field.

\paragraph{Qualitative Observation: Directional Refusal in Qwen2.5}
A distinct behavioral anomaly emerged with Qwen2.5-VL-7B \cite{bai2025qwen25vl}. While the quantitative metrics show a balanced average difference of -7.6, qualitative analysis reveals a systematic failure in instruction following. In a majority of cases under the "Right-to-Left" simple prompt, the model failed to start its description from the right image. Instead, it described the left image first before transitioning to the right, effectively ignoring the directional constraint.

\subsection{Mitigation via Structured Prompting}

Structured prompting (requesting JSON output) significantly reduced bias magnitude in open-source models, though it did not eliminate it. For InternVL2-8B on Caltech data, structured prompting reduced the percentage of strongly left-biased cases from 80\% to 47\%, and the average difference from +472.8 to +33.8 characters.

\subsection{Investigating the Origins of Spatial Bias}

Having established the consistent presence of spatial bias across models and datasets, we investigated three potential sources: language reading direction, explicit dataset annotation instructions, and architectural design choices.

\subsubsection{Language Reading Direction}

The neutral prompting experiment with the Arabic-finetuned AIN model provides crucial evidence. Despite being fine-tuned on right-to-left (RTL) Arabic text, the model described left-positioned content first in 97\% of cases identical to the English-trained Qwen2.5-VL-7B. 

This finding suggests language reading direction is not the primary driver of spatial bias. However, we acknowledge a limitation: AIN is fine-tuned on an existing architecture rather than trained from scratch on RTL languages. While this indicates architectural priors may dominate over fine-tuning, a model pre-trained entirely on RTL data would provide more conclusive evidence.

\subsubsection{Training Data Annotation Patterns}

We examined annotation protocols from two major vision-language datasets: PixMo \cite{deitke2024molmo} and Visual Genome \cite{krishna2017visual}. Neither dataset contains explicit instructions mandating left-to-right description ordering. PixMo annotators are instructed to describe salient objects first through speech-based descriptions, while Visual Genome guidelines emphasize prominent elements without specifying spatial sequences.

This rules out explicit annotation directives as the source. However, implicit human tendencies could still influence training data annotators might naturally describe images left-to-right despite lacking explicit instructions. A quantitative analysis of actual annotations (rather than just guidelines) would be needed to assess this possibility.

\subsubsection{Architectural Factors}

With explicit language and dataset instructions eliminated, we hypothesize architectural origins. Two mechanisms warrant investigation:

\textbf{Positional Embeddings:} The encoding of spatial location for image tokens may introduce systematic processing differences between left and right positions. If positional embeddings (such as RoPE) create asymmetric attention patterns, this could manifest as the observed bias. This hypothesis is consistent with prior work by \cite{zhu2025bias}, which identified unbalanced position embedding design in the language model component as a root cause of spatial inconsistencies.

\textbf{Vision Encoder Design:} Vision transformers process images by dividing them into patches. If patch ordering or spatial relationship encoding favors left-to-right patterns, this architectural choice could produce positional bias independent of language or training data.

\section{Discussion}

\subsection{Need for Mechanistic Understanding}

Our work identifies behavioral manifestations of bias but does not reveal underlying mechanisms. Future research should employ mechanistic interpretability techniques:

\begin{itemize}
\item \textbf{Attention Pattern Analysis:} Visualizing attention weights \cite{vig2019multiscale} across spatial positions during generation could reveal which regions receive preferential processing.
\item \textbf{Activation Patching:} Systematically modifying representations at specific components \cite{wang2023interpretability} could identify which architectural elements are necessary and sufficient for producing spatial bias.
\item \textbf{Intervention Studies:} Testing models with modified positional embeddings or altered patch ordering could causally test the architectural hypothesis.
\end{itemize}

Additionally, quantitative analysis of training data annotations (beyond just guidelines) is needed to measure implicit human left-first tendencies that could influence learned behavior despite the absence of explicit instructions.

\section{Conclusion}

This work demonstrates that Vision-Language Models exhibit systematic spatial attention bias, describing left-positioned content first in 97\% of neutral prompting cases. Through experiments across seven architectures and two datasets, we show this bias persists despite Arabic fine-tuning and explicit spatial instructions, suggesting architectural rather than linguistic origins.

Key findings include: (1) near-universal left-first description preference under neutral prompting, (2) substantial residual bias even under explicit directional instructions, (3) variations in bias magnitude across different visual contexts, and (4) the elimination of language direction and explicit dataset instructions as primary causes.

Achieving position-invariant spatial processing will require explicit architectural interventions informed by mechanistic understanding of how positional embeddings and visual encoders interact during multimodal reasoning. Future VLM development must prioritize interpretability-driven design to identify and address the components responsible for spatial biases.

\section{Limitations and Future Work}

Several limitations should be considered when interpreting these results. Our experiments use horizontally concatenated image pairs from Caltech-101, which contains relatively simple, centered objects. Real-world scenes with complex spatial relationships may exhibit different bias patterns. We tested seven models, but the VLM landscape continues evolving rapidly, and newer architectures may show different characteristics.

Our analysis focuses on behavioral manifestations of bias rather than internal mechanisms producing it. While we propose architectural origins based on elimination of other hypotheses, direct evidence through interpretability studies is needed to confirm these proposals. Future research should investigate mechanistic interpretability studies using attention visualization \cite{vig2019multiscale} and activation patching \cite{wang2023interpretability} to reveal which architectural components produce spatial bias. Additionally, "LLM-as-a-Judge" evaluation pipelines should be utilized to verify semantic adherence to spatial instructions.

\section*{Acknowledgments}

The authors wish to acknowledge the use of \mbox{ChatGPT} in improving the presentation and grammar of the paper. The paper remains an accurate representation of the authors' underlying contributions.

\bibliography{custom}

\end{document}